\documentclass[conference]{IEEEtran}
\IEEEoverridecommandlockouts
\usepackage{cite}
\usepackage{amsmath,amssymb,amsfonts}
\usepackage{algorithmic}
\usepackage{graphicx}
\usepackage{textcomp}
\usepackage{xcolor}
\usepackage[hidelinks]{hyperref}
\usepackage[all]{hypcap} 
\def\BibTeX{{\rm B\kern-.05em{\sc i\kern-.025em b}\kern-.08em
    T\kern-.1667em\lower.7ex\hbox{E}\kern-.125emX}}
\begin{document}

\title{\vspace{0.75cm}%
Bidirectional Optical sensors for Actuation Tracking (BOAT) in soft lattice systems}

\author{\IEEEauthorblockN{1\textsuperscript{st} Petr Trunin}
\IEEEauthorblockA{\textit{Soft BioRobotics and Perception Lab} \\
\textit{Istituto Italiano di Tecnologia (IIT)}\\
\textit{OU ARC at IIT (ARC@IIT)}\\
Genoa, Italy \\
petr.trunin@iit.it}
\and
\IEEEauthorblockN{2\textsuperscript{nd} Carolina Gay}
\IEEEauthorblockA{\textit{Soft BioRobotics and Perception Lab} \\
\textit{Istituto Italiano di Tecnologia (IIT)}\\
Genoa, Italy \\
carolina.gay@iit.it}
\and
\IEEEauthorblockN{3\textsuperscript{rd} Anderson Brazil Nardin}
\IEEEauthorblockA{\textit{Soft BioRobotics and Perception Lab} \\
\textit{Istituto Italiano di Tecnologia (IIT)}\\
Genoa, Italy \\
anderson.nardin@iit.it}
\and
\IEEEauthorblockN{4\textsuperscript{th} Trevor Exley}
\IEEEauthorblockA{\textit{Soft BioRobotics and Perception Lab} \\
\textit{Istituto Italiano di Tecnologia (IIT)}\\
Genoa, Italy \\
trevor.exley@iit.it}
\and
\IEEEauthorblockN{5\textsuperscript{th} Diana Cafiso}
\IEEEauthorblockA{\textit{Soft BioRobotics and Perception Lab} \\
\textit{Istituto Italiano di Tecnologia (IIT)}\\
Genoa, Italy \\
diana.cafiso@iit.it}
\and
\IEEEauthorblockN{6\textsuperscript{th} Lucia Beccai}
\IEEEauthorblockA{\textit{Soft BioRobotics and Perception Lab} \\
\textit{Istituto Italiano di Tecnologia (IIT)}\\
Genoa, Italy \\
lucia.beccai@iit.it}
}

\maketitle

\begin{abstract}
The growing adoption of lattice-based structures in soft robotics creates a need for advanced sensing solutions capable of monitoring their global deformation, particularly compression and extension. In this work, we address this challenge by introducing a novel optical sensor based on two patterned waveguides arranged in an ellipsoidal geometry. This Bidirectional Optical sensor for Actuation Tracking (BOAT) is seamlessly co-printed with a lattice structure actuated by an embedded pneumatic artificial muscle (PAM), and its performance is assessed. During PAM elongation or contraction, the bending of the embedded BOAT waveguides induces output signal variations that enable a clear discrimination between compression and extension states.

The designs of, both, each specific waveguide structure (by surface patterning), and of the sensorized lattice-based unit embedding two BOATs, are supported by numerical simulations. Experimental calibration over 100 consecutive pressure cycles ranging from +50 kPa to $-$40 kPa demonstrate a highly repeatable response, allowing a reliable distinction between extension and compression.

Finally, sensor feedback  is used to implement a digital shadow, enabling continuous synchronization between the whole sensorized unit and its virtual counterpart. These results establish BOAT as a powerful and reliable approach for deformation monitoring in soft lattice-based robotic systems.
\end{abstract}


\section{Introduction}
In the past decade, soft robotic systems have exhibited increasing structural and functional complexity, driven by the diversification of application domains and advances in additive manufacturing technologies\cite{oh_architected_2026}. This way the soft robotic approach is representing a sound alternative to mechatronics to build intelligent systems where the integration among sensing, actuation, robot body and control is key. Among emerging architectural strategies, three-dimensional printed lattice-based robots have gained particular attention\cite{guan_lattice_2025, guan_trimmed_2023}. 
\begin{figure}[!htb]
\centering
  \includegraphics[width=1\columnwidth]{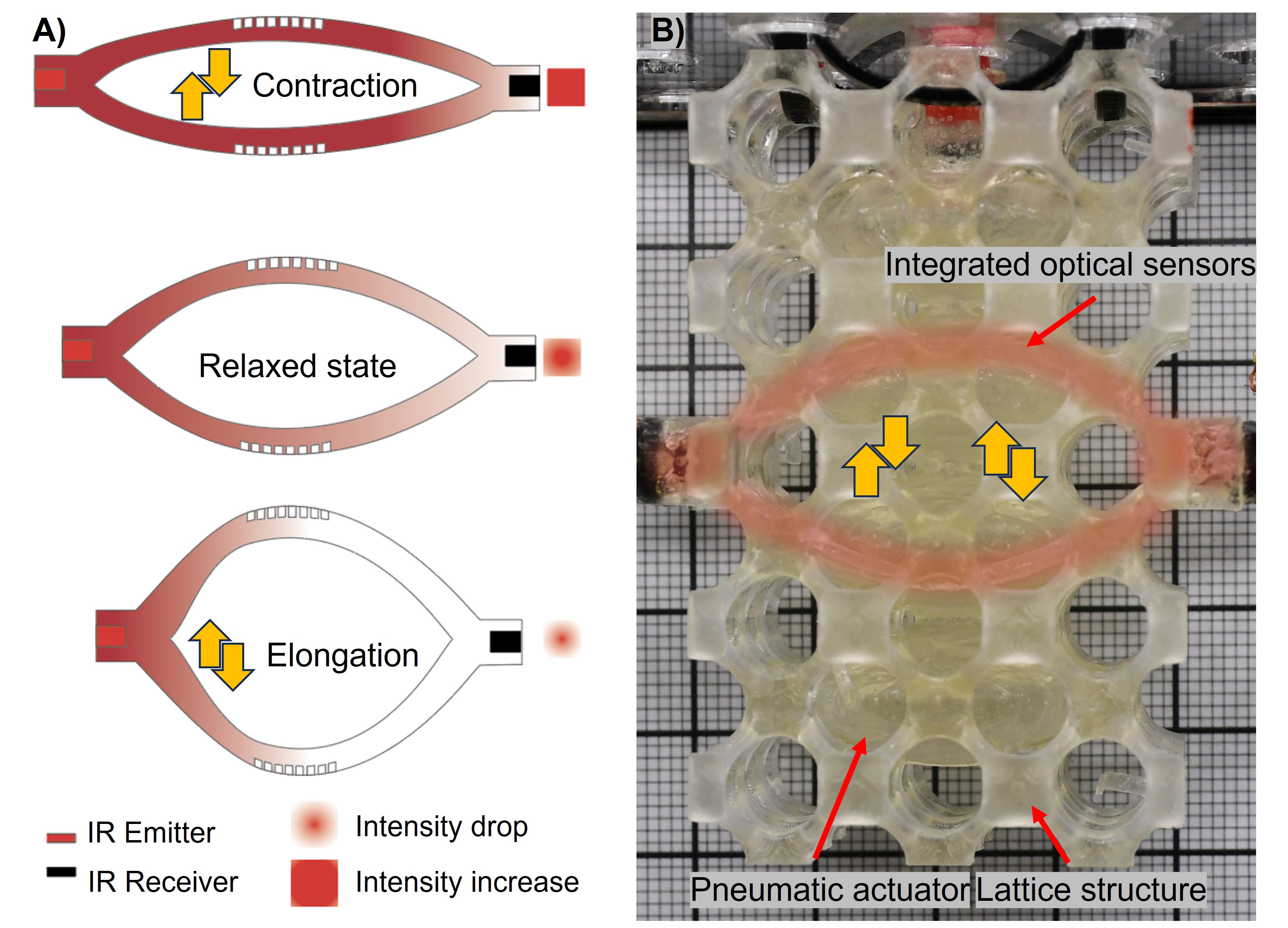}
  \caption{A. Representation of the BOAT working principle. B. Frontal view of the testing sample, consisting of a lattice structure embedding two BOATs integrated and a pneumatic artificial muscle (PAM).}
  \label{fig:figure1}
\end{figure}
Lattice architectures enable tunable mechanical behavior through parametric control of unit-cell geometry, strut thickness, and topology, thereby allowing systematic modulation of global stiffness, compliance, and anisotropy \cite{oh_architected_2026}. This structural programmability supports adaptable and application-specific design. In addition, lattice-based systems are compatible with multiple actuation principles, including tendon-driven mechanisms\cite{guan_lattice_2025}, pneumatic actuation\cite{joe_jointless_2023}, and shape-memory alloys\cite{chen_programmable_2026}, which broadens their operational scope.

Despite these advantages, sensor integration within lattice-based robots remains insufficiently explored. Usually, proprioceptive sensing is achieved by bonding strain sensors to regions of maximum curvature \cite{wang_soft_2024, wang_toward_2018, yang_computational_2024, so_shape_2021}, while tactile sensing is implemented via pressure sensors attached to surfaces that make contact with the exterior \cite{roberts_soft_2021, qu_recent_2023, zhao_optoelectronically_2016}. Such surface-mounted strategies rely on the presence of continuous, regular geometries. In contrast, lattice-based architectures lack extended planar surfaces, and their internal deformation fields are spatially distributed and structurally heterogeneous. As a result, sensing elements must be embedded within the structure at carefully selected locations to ensure meaningful signal acquisition without compromising mechanical performance.

Previous demonstrations have addressed tactile sensing and bending detection in lattice-integrated systems \cite{trunin_melegros_2026, exley_monolithic_2025}. However, sensorizing both dominant deformation modes in many lattice-based actuators, i.e. elongation and compression, remains a challenge. Due to the intrinsic geometry of lattice structures, global motion is generated by localized strut compression or extension. These deformation modes are challenging to monitor using conventional sensing approaches, as strain distribution is non-uniform and geometrically complex.

In this work, we present the implementation of a Bidirectional Optical sensor for Actuation Tracking (BOAT) based on a bending-induced\cite{trunin_design_2025} intensity modulation principle for detecting compression and elongation in a lattice-integrated PAM (Figure~\ref{fig:figure1}). The sensing element comprises two patterned optical waveguides arranged in an ellipsoidal configuration (Figure~\ref{fig:figure1}A). Under lattice compression, the waveguides tend toward a straighter configuration, reducing optical loss and increasing the light intensity measured at the receiver. Conversely, during lattice elongation, the waveguides undergo increased bending, resulting in higher optical attenuation and a corresponding decrease in received intensity. The sensor is fabricated concurrently with the pneumatic chambers and integrated directly into the lattice structure to validate the sensing principle. With respect to our previous work on monolithic optical waveguides~\cite{trunin_design_2025} and
lattice-integrated soft actuators~\cite{trunin_melegros_2026, exley_monolithic_2025}, the specific contributions of this paper are: (i)~a bidirectional sensing element based on an ellipsoidal twin-waveguide geometry that discriminates compression from elongation within a single actuated unit; (ii)~a coupled mechanical--optical co-design pipeline to define the waveguide surface patterning directly from the simulated deformation states of the actuator; and (iii)~a sensor-driven digital shadow that mirrors the PAM in real time.

\section{Methods and Experimental Validation}

\subsection{Design and Mechanical Simulation}

The system consists of a bladder-like PAM fully embedded within a compliant lattice envelope and integrated optical waveguide sensors (Figure~\ref{fig:figure1}). The PAM geometry follows the bladder architecture previously reported in \cite{trunin_melegros_2026}. The spacing between adjacent bladders is defined to be equal to the lattice unit-cell size, ensuring geometric coupling between the actuator and the surrounding lattice during pressurization. The overall geometry is created parametrically in Grasshopper (Rhinoceros 3D) using the lattice-generation workflow introduced in \cite{exley_monolithic_2025}.

The lattice corresponds to a quasi-IWP morphology with a unit-cell size of 12.5 mm and a minimum strut thickness of 1.5 mm, selected based on manufacturability constraints imposed by the optical waveguide surface patterning. The effective Young’s modulus used for simulation is obtained experimentally from uniaxial compression tests of this lattice configuration (18.34 ± 1.86 kPa, N = 3).

\begin{figure*}[!htb]
\centering
  \includegraphics[width=0.8\textwidth]{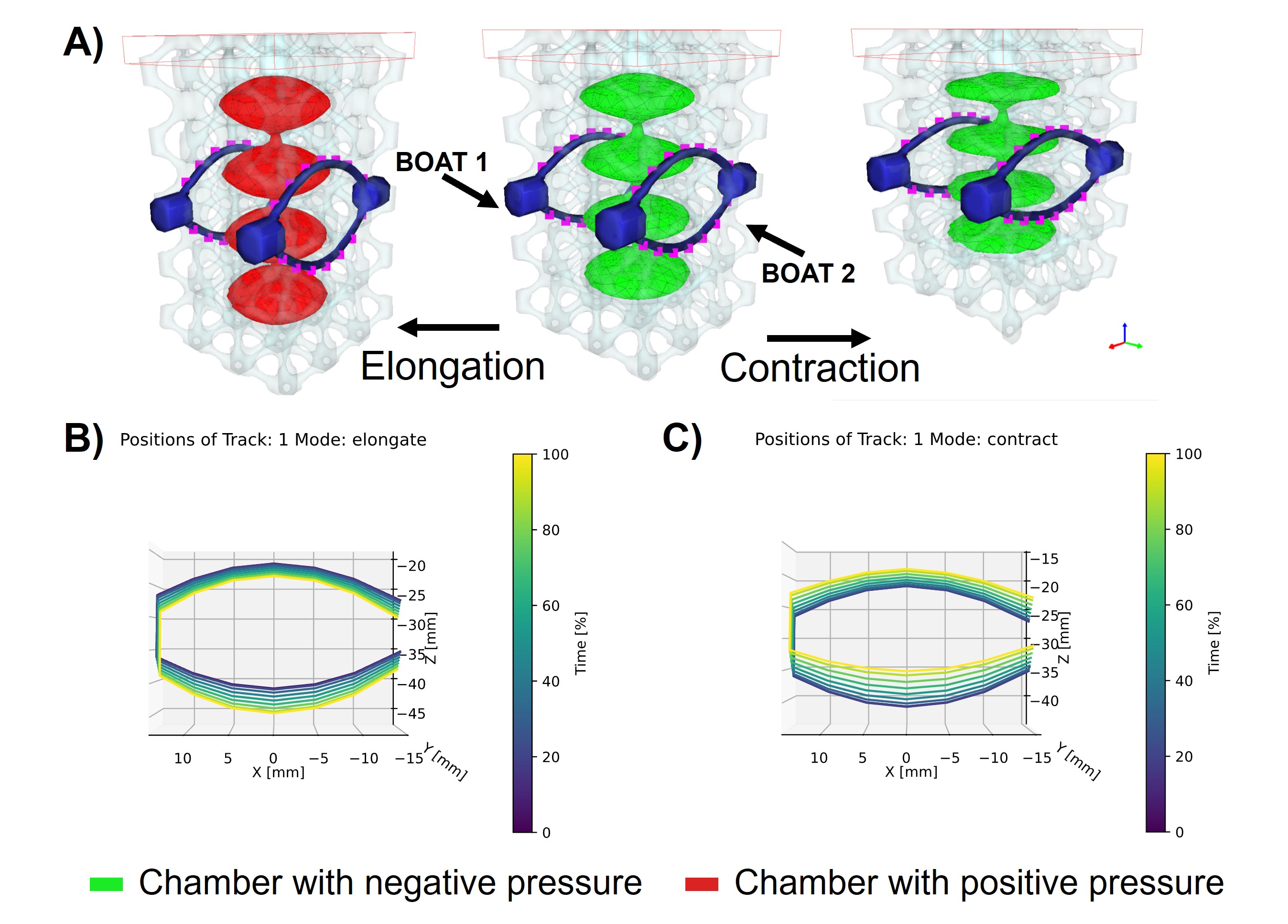}
  \caption{A. Configuration of the sensorized actuator from the SOFA simulation at three deformation states: elongation (50 kPa), rest, and compression ($-$50 kPa). The BOATs are shown in blue, the PAM in red and green, and the tracked points in pink. B,C. Deformation curve profiles retrieved from the tracked points extracted from simulations during compression and elongation.}
  \label{fig:figure2}
\end{figure*}

The BOAT consists of two patterned optical waveguides arranged in an ellipsoidal configuration and embedded across the actuator length. Each waveguide is generated by selecting horizontally aligned anchor points at the lattice nodes external to the PAM. These points are then interpolated to define a continuous centerline, and the waveguide cross-section is built along the resulting curve. Points are sampled along each centerline at 0.5 mm intervals and used as tracking points in the mechanical simulation.

Mechanical behavior is evaluated in SOFA (Simulation Open Framework Architecture)~\cite{duriez_realistic_2006}\cite{faure_sofa_2012} using a de-featured geometry with a homogenized representation of the lattice \cite{nardin_exploring_2025, exley_monolithic_2025}. Pneumatic pressure is prescribed on the internal cavity surfaces and varied from $-$50 kPa to 50 kPa to reproduce the full actuation cycle, spanning elongation and compression (Figure~\ref{fig:figure2}A). During simulation, the tracked points are recorded at each time step to extract the deformation curves of the waveguides (Figure~\ref{fig:figure2}B,C). These curve profiles provide the mechanical input for the optical modeling stage, where the waveguide surface patterns are optimized.

\subsection{Optical simulation}

The working principle of soft optical sensors relies on the reduction of light intensity at the receiver during bending, caused by increased scattering of optical rays at the curved waveguide surface. As the waveguide deforms, light propagation is perturbed, leading to higher optical losses \cite{wang_soft_2024}.
The addition of a surface pattern, which consists of a series of rectangular cavities,  enhances light refraction and increases surface scattering along the waveguide, resulting in improved sensor sensitivity \cite{trunin_design_2025}.

To evaluate the proposed optical sensor behavior and optimize the geometrical parameters of the patterns (rectangular cavities) at the waveguides' surface, finite element simulations were performed using COMSOL Multiphysics® (COMSOL Inc., Sweden). To reduce computational cost, a two-dimensional (2D) model was adopted, coupling the Ray Optics module through the Multiphysics interface.

\begin{figure*}[!htb]
\centering
  \includegraphics[width=0.8\textwidth]{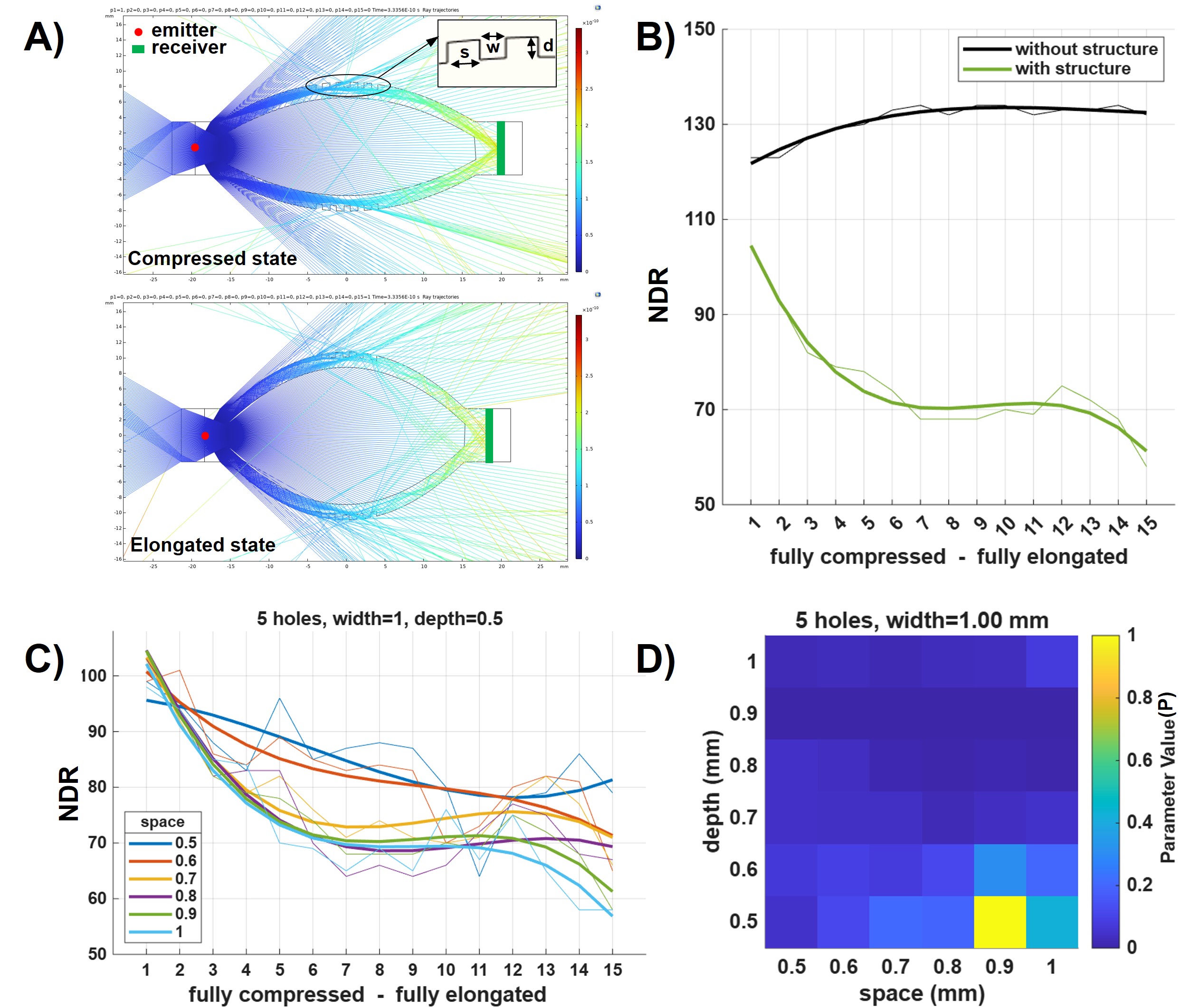}
  \caption{ A. Visualization of optical simulation of the BOAT with superficial pattern (5 cavities, width=1 mm depth=0.5 mm, space=0.9 mm. Two deformation states are shown, the maximum compression and the maximum elongation.   B. Number of detected rays (NDR) and approximated third-degree polynomial (aNDR) along the deformation. C. All the combination of the parameter space (with 5 cavities, width=1 mm depth= 0.5 mm) are represented as number of detected rays (NDR) and approximated third-degree polynomial (aNDR) along the deformation. D. Color map showing the Parameter \textit{P} (normalized with respect to its maximum value) for all combination of depth and space with 5 cavities, width =  1 mm.}
  \label{fig:figure3}
\end{figure*}

Fourteen-point coordinates (plane x-z, 7 for each waveguide) along the two sensors' curves (in pink in Figure~\ref{fig:figure2}A) are extracted from SOFA simulations with a sampling frequency of 10 Hz. The data is first processed in MATLAB and then imported into COMSOL. Specifically, after oversampling, ten points for each waveguide are selected for each curve and implemented using the Interpolation Curve feature to reconstruct the continuous profile.
In total, fifteen curves (each defined by ten points) are considered for each of the two embedded waveguides: seven under compression, seven under elongation, and one corresponding to the rest position. 
BOAT is then reconstructed in the Geometry module using the actual dimensions and parametrized across the fifteen deformation states, from maximum compression to maximum elongation (Figure~\ref{fig:figure3}A). 
Within the Ray Optics module, the emitter is modeled as a point source emitting 250 rays with a conical distribution and an aperture angle of 120°, in accordance with the emitter specifications. It is positioned 3 mm from the junction point.
The receiver, defined as a segment, is placed 3 mm from the opposite junction point. The light wavelength is set to 860 nm, and the maximum number of secondary rays is set to 100.
The mesh is generated automatically using the “finer” element size setting.

The desired outcome is to maximize the number of detected rays under compression and minimize it under elongation. To enhance this contrast, a pattern is introduced along the waveguides, as a series of rectangular cavities centered along the midline of each waveguide profile.  
Three geometric parameters are varied from 0.5 to 1.0 mm with a 0.1 mm step: cavity width, cavity depth, and spacing between adjacent cavities (space in Figure~\ref{fig:figure3}A). 
The optical simulations enable evaluation of BOAT's response across all parameter combinations (6 × 6 × 6) and the 15 deformation states, resulting in a total of 3,240 simulations (around 8 seconds each).
Moreover, this calculation is performed with different number of cavities: 3, 5 and 7.

All simulation results are exported and processed in MATLAB to identify the optimal parameter combination. For each configuration, the number of detected rays (NDR) is plotted across the fifteen deformation states, and its evolution was approximated using a third-degree polynomial fit (aNDR) (Figure~\ref{fig:figure3}B,C). To quantitatively determine the best parameter set, the following metric is defined:
\begin{equation}
P = \frac{\Delta}{n_{sign} \cdot RMSE} 
\end{equation}
where $\Delta$ represents the difference between the fitted aNDR at maximum compression and at maximum elongation. A larger $\Delta$ indicates higher sensor sensitivity. The denominator includes $n_{sign}$ and $RMSE$. The term $n_{sign}$ represents the number of instances in which the NDR increases during the transition from the compressed to the elongated state. Ideally, the detected rays count should decrease monotonically  during elongation; therefore, any increase is penalized. $RMSE$ denotes the root mean squared error between the fitted curve (aNDR) and the simulated NDR values, accounting for the goodness of fit. \textit{P} is therefore a composite figure of merit that simultaneously maximizes sensitivity, monotonicity, and linearity. 
This parameter \textit{P} is calculated for all simulations and normalized with the global maximum.  
The result is shown in Figure~\ref{fig:figure3}B,D.
\subsection{Calibration}

\begin{figure*}[!htb]
\centering
  \includegraphics[width=0.75\textwidth]{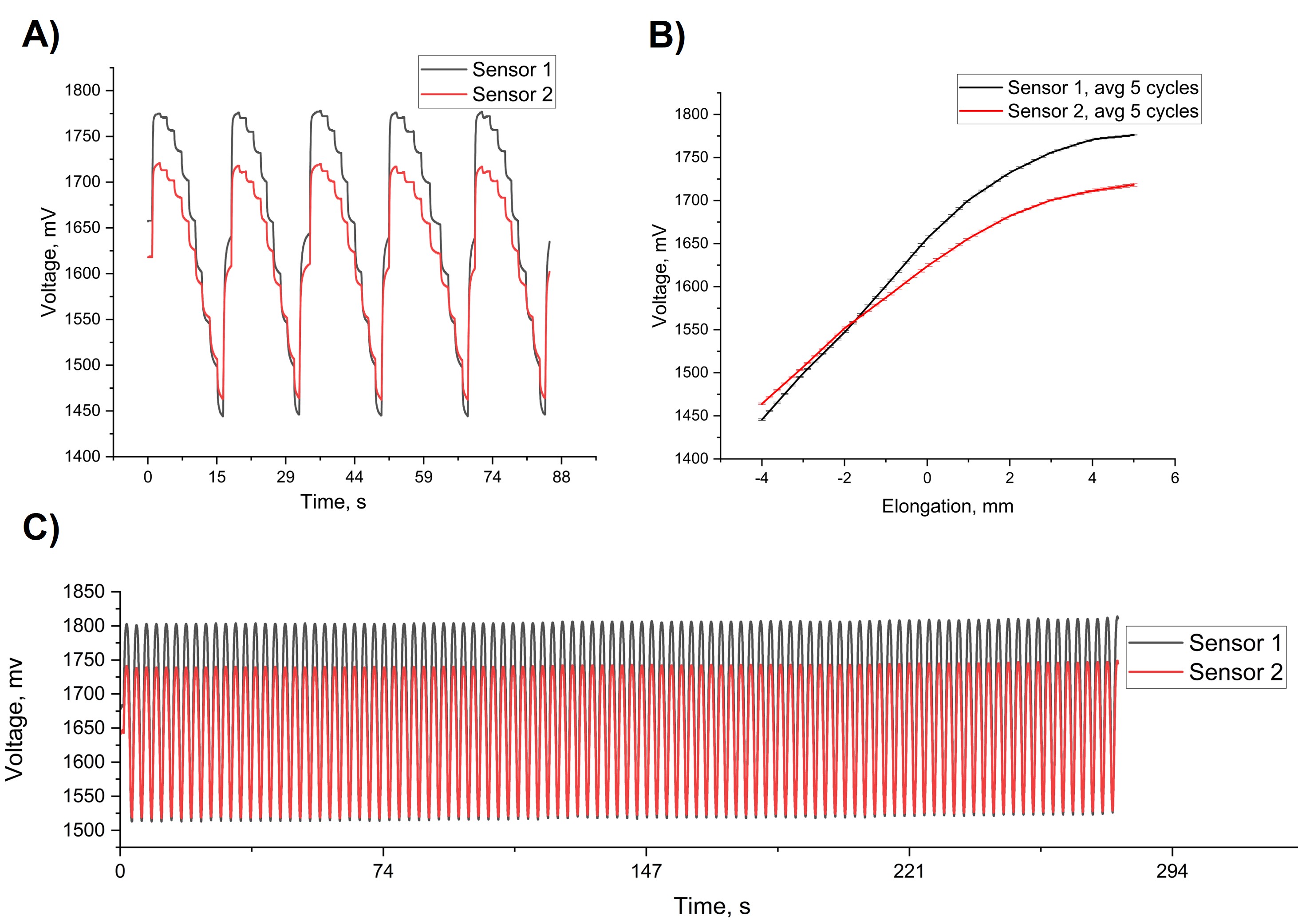}
  \caption{A. Graph of the signals from the two BOATs of a single sensorized actuator (SA) during five cycles of pressure variation between +50 kPa and $-$40 kPa (10 kPa steps). Each cycle started at the elongation state (+50 kPa) and ended at the compression state ($-$40 kPa). B Representation of the calibration functions of the sensorized actuator SA retrieved from the five cycle test. The x-axis indicates the actuator deformation (mm), and the y-axis shows the responses of the two sensors (V). C. SA Sensors’ responses under continuous pressure variations between +50 kPa and $-$40 kPa over 100 cycles.}
  \label{fig:figure4}
\end{figure*}

After determining the optimal number of cavities in the sensor design, the parameters are transferred to Grasshopper for the final design stage. The prototype uses $n=5$ cavities with width $w=1.0$\,mm, depth $d=0.5$\,mm, and spacing $s=0.9$\,mm, identified by the maximum of $P$ in Fig.~\ref{fig:figure3}D. Based on this configuration, three sensorized actuators (SA) are fabricated via SLA 3D printing (Form~4, Formlabs, USA) using Elastic 50A resin. The optical chain consists of an 860\,nm IR emitter (VSMY1850, Vishay) and a matched photodiode receiver (TEMT7100X01, Vishay), read at 100\,Hz by a microcontroller (CY8C5667AXI-LP040) supplied at 3.3\,V. To minimize the influence of external illumination, the acquisition board is programmed to operate with sequential activation of the emitters. When an emitter is deactivated, the receiver measures the ambient light level, which is then subtracted from the signal acquired during the active phase, providing an effective rejection of background illumination without additional optical insulation. Each actuator integrates two BOATs (hearafter Sensor 1 and Sensor 2) and designated cavities for electronic components, as illustrated in Figure~\ref{fig:figure1}B. A photoemitter and photoreceiver are inserted into the corresponding housings and secured with small amounts of uncured Elastic 50A, followed by UV curing for 30 seconds. This step ensures material continuity within the monolithic structure and mechanically stabilizes the embedded electronic components.

Following fabrication, each SA is mounted on a plexiglass support and connected to a printed circuit board (PCB) and pneumatic pumps. The pressure input is controlled within a range from $-$40 kPa to +50 kPa. The lower limit is defined by the onset of buckling in the lattice structure below $-$40 kPa, which alters the mechanical response and consequently affects sensor performance. The actuator is positioned in front of a millimetric reference grid to measure the displacement of its distal tip. Pressure is varied from +50 kPa to $-$40 kPa in 10 kPa increments over five loading-unloading cycles, while raw signals from both embedded sensors are recorded (Figure~\ref{fig:figure4}A). The data is segmented into individual cycles and plotted against measured elongation to generate calibration curves (Figure~\ref{fig:figure4}B). 
The calibration functions for each sensor are obtained using a third-order polynomial regression.

For Sensor~1:
\begin{equation}
y_1(x) = 1651.9842 + 46.4413\,x - 2.6934\,x^2 - 0.36251\,x^3 ,
\end{equation}
with $R^2 = 0.99971$.

For Sensor~2:
\begin{equation}
y_2(x) = 1624.5567 + 33.37911\,x - 2.3102\,x^2 - 0.14139\,x^3 ,
\end{equation}
with $R^2 = 0.99991$.

The effective sensitivity of the resulting sensing system, obtained from the averaged calibration of both sensors, is

\begin{equation}
S_{\mathrm{sys}} = 34.2 \ \mathrm{mV/mm}.
\end{equation}

The offset between Sensor 1 and Sensor 2 reflects unit-to-unit variability in cavity printing and emitter/receiver placement within the embedded housings; for this reason, each fabricated actuator should be individually calibrated, while the calibration procedure itself is unchanged. To evaluate repeatability and structural stability, 100 consecutive pressure cycles between +50 kPa and $-$40 kPa are applied (Figure~\ref{fig:figure3}C). Based on the calibration results, a calibration function is derived to establish a digital representation (“digital shadow”) of the sensorized actuator.

\subsection{Real-time tests and digital shadow}

Following calibration, real-time experiments were conducted to validate the sensor’s performance. The calibration functions for the two SA sensors are implemented in the serial port reader and used as inputs to the simulation environment. Because the simulation is displacement-driven, the measured voltages are converted into displacement values through the calibration functions. The computed displacements are then visualized in real time at approximately 2 frames per second (Video S1). Figure 5A and 5B present representative simulation states corresponding to the real-time configuration of the SA shown in Figure 5C and 5D.

\begin{figure}[!htb]
\centering
  \includegraphics[width=1\columnwidth]{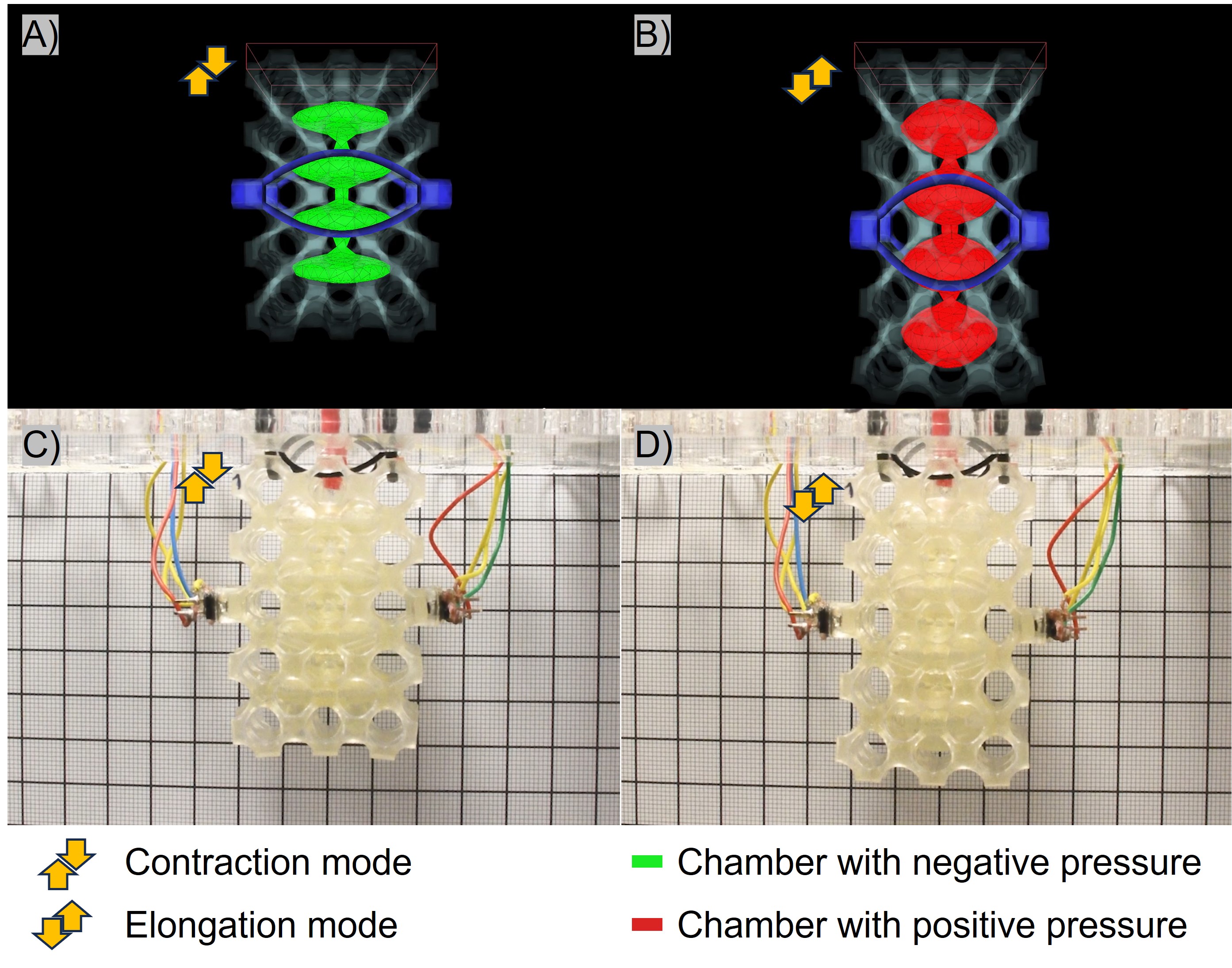}
  \caption{ A,B. Representative states of the SOFA simulation driven in real time by calibrated sensors data. C,D. Corresponding real-time configurations of the SA during experimental actuation. The simulation reproduces the instantaneous mechanical state inferred from the measured displacement.}
  \label{fig:figure5}
\end{figure}

The SOFA simulation runs at approximately 2 frames per second (FPS) on a workstation equipped with an Intel Core Ultra 9 185H processor (16 physical cores, 22 logical processors, base frequency 2.30 GHz) and 32 GB of RAM. This configuration enables real-time monitoring of selected kinematic observables, such as the position of tracked points sampled along the actuator and waveguide centerlines. The computational performance is therefore sufficient to support online comparison between simulated and experimentally measured quantities.

In its current implementation, the framework constitutes a digital shadow. Sensor data acquired from the PAM is used to update the digital model state. The information flow is unidirectional, from the physical system to its digital representation. The simulation, therefore, reflects the instantaneous mechanical configuration inferred from experimental inputs, but does not yet impose corrective actions on the hardware.

The primary benefit of this digital shadow architecture lies in its capacity to provide an idealized reference behavior against which the physical system can be continuously compared. Because the simulation tracks the full displacement of the actuator, including the position of each mesh node, it defines a nominal configuration for any prescribed pressure. For instance, at a given internal pressure, the model predicts a specific tip displacement. The corresponding experimentally measured tip position can be directly compared, yielding a deviation.

Systematic growth of this deviation over repeated actuation cycles may indicate progressive material aging, stiffness degradation, or entry into a plastic regime not captured by the nominal elastic model. Furthermore, if the applied pressure reported by the physical system does not produce the deformation predicted by the model, and if this discrepancy increases over time or under sustained loading, it may suggest fluid leakage or compromised sealing. Because the simulation assumes ideal boundary conditions and material integrity, any persistent mismatch between digital and physical states can be interpreted as a diagnostic signal.

Transitioning toward a full digital twin would require closing the loop between the digital and physical domains. In such an architecture, the deviation would not only be monitored but actively used to compute corrective actions. 

\section{Conclusion}

In this work, a novel design for optical sensors capable of measuring both compression and elongation was introduced. The sensing mechanism relies on bending behavior combined with superficial patterning to modulate the optical response under deformation.

The proposed simulation-driven methodology enables a direct and reliable manufacturing workflow without iterative trial-and-error adjustments. The process is grounded in coupled mechanical and optical simulations of the complete structure. Once the virtual model is validated, the sensorized actuator can be fabricated in a calibration-ready configuration, reducing post-processing adjustments. Validation here is restricted to a single material (Elastic~50A) and a single lattice geometry; the simulation-driven design pipeline itself is, however, geometry- and material-agnostic, and can be applied to other actuators provided that the underlying mechanical and optical models are re-parameterized.

Experimental calibration demonstrated a repeatable and low-noise signal, indicating stable transduction characteristics and consistent performance across actuation cycles. The hysteresis for the sensors over 100 cycles is less than 5\%: 4.95\% for Sensor 1 and 4.46\% for Sensor 2. The sensitivity is 39.3 mV/mm and 29.2 mV/mm, respectively. Signal-to-noise ratio is 44.7dB and 45.2dB. These results support the suitability of the proposed sensor architecture for future applications requiring reliable deformation monitoring.

Following calibration, a digital shadow was implemented. Sensor feedback, processed through the calibration functions, was used to drive a real-time SOFA simulation that reproduces the actuator’s mechanical state. This configuration enables continuous synchronization between the PAM and its virtual representation. Future work will extend this architecture toward a full digital twin by introducing bidirectional interaction and closed-loop control.

\bibliographystyle{IEEEtran}

\bibliography{references}

@article{zhao_optoelectronically_2016,
	title = {Optoelectronically innervated soft prosthetic hand via stretchable optical waveguides},
	volume = {1},
	issn = {2470-9476},
	url = {https://www.science.org/doi/10.1126/scirobotics.aai7529},
	doi = {10.1126/scirobotics.aai7529},
	language = {en},
	number = {1},
	urldate = {2026-02-19},
	journal = {Science Robotics},
	author = {Zhao, Huichan and O’Brien, Kevin and Li, Shuo and Shepherd, Robert F.},
	month = dec,
	year = {2016},
	pages = {eaai7529},
}

@article{qu_recent_2023,
	title = {Recent {Progress} in {Advanced} {Tactile} {Sensing} {Technologies} for {Soft} {Grippers}},
	volume = {33},
	issn = {1616-301X, 1616-3028},
	url = {https://advanced.onlinelibrary.wiley.com/doi/10.1002/adfm.202306249},
	doi = {10.1002/adfm.202306249},
	language = {en},
	number = {41},
	urldate = {2026-02-19},
	journal = {Advanced Functional Materials},
	author = {Qu, Juntian and Mao, Baijin and Li, Zhenkun and Xu, Yining and Zhou, Kunyu and Cao, Xiangyu and Fan, Qigao and Xu, Minyi and Liang, Bin and Liu, Houde and Wang, Xueqian and Wang, Xiaohao},
	month = oct,
	year = {2023},
	pages = {2306249},
}

@article{roberts_soft_2021,
	title = {Soft {Tactile} {Sensing} {Skins} for {Robotics}},
	volume = {2},
	issn = {2662-4087},
	url = {https://doi.org/10.1007/s43154-021-00065-2},
	doi = {10.1007/s43154-021-00065-2},
	language = {en},
	number = {3},
	urldate = {2026-02-19},
	journal = {Current Robotics Reports},
	author = {Roberts, Peter and Zadan, Mason and Majidi, Carmel},
	month = sep,
	year = {2021},
	pages = {343--354},
}

@article{yang_computational_2024,
	title = {Computational design of ultra-robust strain sensors for soft robot perception and autonomy},
	volume = {15},
	issn = {2041-1723},
	url = {https://www.nature.com/articles/s41467-024-45786-y},
	doi = {10.1038/s41467-024-45786-y},
	language = {en},
	number = {1},
	urldate = {2026-02-19},
	journal = {Nature Communications},
	author = {Yang, Haitao and Ding, Shuo and Wang, Jiahao and Sun, Shuo and Swaminathan, Ruphan and Ng, Serene Wen Ling and Pan, Xinglong and Ho, Ghim Wei},
	month = feb,
	year = {2024},
	pages = {1636},
}

@article{chen_programmable_2026,
	title = {Programmable structure with shape memory materials for soft robotics},
	volume = {35},
	issn = {0964-1726, 1361-665X},
	url = {https://iopscience.iop.org/article/10.1088/1361-665X/ae2a85},
	doi = {10.1088/1361-665X/ae2a85},
	number = {1},
	urldate = {2026-02-19},
	journal = {Smart Materials and Structures},
	author = {Chen, Qianyi and Wu, Ruochen and Schott, Dingena and Jovanova, Jovana},
	month = jan,
	year = {2026},
	pages = {015049},
}

@article{guan_trimmed_2023,
	title = {Trimmed helicoids: an architectured soft structure yielding soft robots with high precision, large workspace, and compliant interactions},
	volume = {1},
	issn = {2731-4278},
	shorttitle = {Trimmed helicoids},
	url = {https://www.nature.com/articles/s44182-023-00004-7},
	doi = {10.1038/s44182-023-00004-7},
	language = {en},
	number = {1},
	urldate = {2026-02-19},
	journal = {npj Robotics},
	author = {Guan, Qinghua and Stella, Francesco and Della Santina, Cosimo and Leng, Jinsong and Hughes, Josie},
	month = oct,
	year = {2023},
	pages = {4},
}

@article{oh_architected_2026,
	title = {Architected {Materials} for {Soft} {Robotics}},
	issn = {0884-2914, 2044-5326},
	url = {https://link.springer.com/10.1557/s43578-025-01778-2},
	doi = {10.1557/s43578-025-01778-2},
	language = {en},
	urldate = {2026-02-19},
	journal = {Journal of Materials Research},
	author = {Oh, EunBi and Kim, Taekyoung and Kaarthik, Pranav and Truby, Ryan L.},
	month = jan,
	year = {2026},
}

@incollection{faure_sofa_2012,
	address = {Berlin, Heidelberg},
	title = {{SOFA}: {A} {Multi}-{Model} {Framework} for {Interactive} {Physical} {Simulation}},
	isbn = {9783642290145},
	shorttitle = {{SOFA}},
	url = {https://doi.org/10.1007/8415_2012_125},
	language = {en},
	urldate = {2026-02-19},
	booktitle = {Soft {Tissue} {Biomechanical} {Modeling} for {Computer} {Assisted} {Surgery}},
	publisher = {Springer},
	author = {Faure, François and Duriez, Christian and Delingette, Hervé and Allard, Jérémie and Gilles, Benjamin and Marchesseau, Stéphanie and Talbot, Hugo and Courtecuisse, Hadrien and Bousquet, Guillaume and Peterlik, Igor and Cotin, Stéphane},
	editor = {Payan, Yohan},
	year = {2012},
	doi = {10.1007/8415_2012_125},
	pages = {283--321},
}

@article{trunin_design_2025,
	title = {Design and {3D} printing of soft optical waveguides towards monolithic perceptive systems},
	volume = {100},
	issn = {22148604},
	url = {https://linkinghub.elsevier.com/retrieve/pii/S221486042500051X},
	doi = {10.1016/j.addma.2025.104687},
	language = {en},
	urldate = {2026-02-19},
	journal = {Additive Manufacturing},
	author = {Trunin, Petr and Cafiso, Diana and Beccai, Lucia},
	month = feb,
	year = {2025},
	pages = {104687},
}

@article{wang_soft_2024,
	title = {Soft {Optical} {Waveguides} for {Biomedical} {Applications}, {Wearable} {Devices}, and {Soft} {Robotics}: {A} {Review}},
	volume = {6},
	issn = {2640-4567, 2640-4567},
	shorttitle = {Soft {Optical} {Waveguides} for {Biomedical} {Applications}, {Wearable} {Devices}, and {Soft} {Robotics}},
	url = {https://advanced.onlinelibrary.wiley.com/doi/10.1002/aisy.202300482},
	doi = {10.1002/aisy.202300482},
	language = {en},
	number = {1},
	urldate = {2026-02-19},
	journal = {Advanced Intelligent Systems},
	author = {Wang, Xuechun and Li, Zilong and Su, Lei},
	month = jan,
	year = {2024},
	pages = {2300482},
}

@article{trunin_melegros_2026,
	title = {{MELEGROS}: {Monolithic} {Elephant}‐{Inspired} {Gripper} with {Optical} {Sensors}},
	issn = {2198-3844, 2198-3844},
	shorttitle = {{MELEGROS}},
	url = {https://advanced.onlinelibrary.wiley.com/doi/10.1002/advs.202518878},
	doi = {10.1002/advs.202518878},
	language = {en},
	urldate = {2026-02-19},
	journal = {Advanced Science},
	author = {Trunin, Petr and Cafiso, Diana and Nardin, Anderson Brazil and Exley, Trevor and Beccai, Lucia},
	month = feb,
	year = {2026},
	pages = {e18878},
}

@misc{exley_monolithic_2025,
	title = {Monolithic {Units}: {Actuation}, {Sensing}, and {Simulation} for {Integrated} {Soft} {Robot} {Design}},
	copyright = {arXiv.org perpetual, non-exclusive license},
	shorttitle = {Monolithic {Units}},
	url = {https://arxiv.org/abs/2511.13120},
	doi = {10.48550/ARXIV.2511.13120},
	urldate = {2026-02-19},
	publisher = {arXiv},
	author = {Exley, Trevor and Nardin, Anderson Brazil and Trunin, Petr and Cafiso, Diana and Beccai, Lucia},
	year = {2025},
}

@article{duriez_realistic_2006,
	title = {Realistic haptic rendering of interacting deformable objects in virtual environments},
	volume = {12},
	copyright = {https://ieeexplore.ieee.org/Xplorehelp/downloads/license-information/IEEE.html},
	issn = {1077-2626},
	url = {http://ieeexplore.ieee.org/document/1541998/},
	doi = {10.1109/TVCG.2006.13},
	number = {1},
	urldate = {2026-02-19},
	journal = {IEEE Transactions on Visualization and Computer Graphics},
	author = {Duriez, C. and Dubois, F. and Kheddar, A. and Andriot, C.},
	month = jan,
	year = {2006},
	pages = {36--47},
}

@inproceedings{nardin_exploring_2025,
	address = {Lausanne, Switzerland},
	title = {Exploring {Effective} {Approaches} to the {Modelling} of {Lattice}-based {Structures} for {Soft} {Robots}},
	copyright = {https://doi.org/10.15223/policy-029},
	isbn = {9798331520205},
	url = {https://ieeexplore.ieee.org/document/11020867/},
	doi = {10.1109/RoboSoft63089.2025.11020867},
	urldate = {2026-02-19},
	booktitle = {2025 {IEEE} 8th {International} {Conference} on {Soft} {Robotics} ({RoboSoft})},
	publisher = {IEEE},
	author = {Nardin, Anderson B. and Joe, Seonggun and Bliah, Ouriel and Magdassi, Shlomo and Beccai, Lucia},
	month = apr,
	year = {2025},
	pages = {1--6},
}

@article{joe_jointless_2023,
	title = {Jointless {Bioinspired} {Soft} {Robotics} by {Harnessing} {Micro} and {Macroporosity}},
	volume = {10},
	issn = {2198-3844, 2198-3844},
	url = {https://onlinelibrary.wiley.com/doi/10.1002/advs.202302080},
	doi = {10.1002/advs.202302080},
	language = {en},
	number = {23},
	urldate = {2025-09-24},
	journal = {Advanced Science},
	author = {Joe, Seonggun and Bliah, Ouriel and Magdassi, Shlomo and Beccai, Lucia},
	month = aug,
	year = {2023},
	pages = {2302080},
}

@article{so_shape_2021,
	title = {Shape {Estimation} of {Soft} {Manipulator} {Using} {Stretchable} {Sensor}},
	volume = {2021},
	copyright = {http://creativecommons.org/licenses/by/4.0/},
	issn = {2692-7632},
	url = {https://spj.science.org/doi/10.34133/2021/9843894},
	doi = {10.34133/2021/9843894},
	language = {en},
	urldate = {2025-08-25},
	journal = {Cyborg and Bionic Systems},
	author = {So, Jinho and Kim, Uikyum and Kim, Yong Bum and Seok, Dong-Yeop and Yang, Sang Yul and Kim, Kihyeon and Park, Jae Hyeong and Hwang, Seong Tak and Gong, Young Jin and Choi, Hyouk Ryeol},
	month = jan,
	year = {2021},
	pages = {2021/9843894},
}

@article{wang_toward_2018,
	title = {Toward {Perceptive} {Soft} {Robots}: {Progress} and {Challenges}},
	volume = {5},
	issn = {2198-3844, 2198-3844},
	shorttitle = {Toward {Perceptive} {Soft} {Robots}},
	url = {https://onlinelibrary.wiley.com/doi/10.1002/advs.201800541},
	doi = {10.1002/advs.201800541},
	language = {en},
	number = {9},
	urldate = {2025-08-25},
	journal = {Advanced Science},
	author = {Wang, Hongbo and Totaro, Massimo and Beccai, Lucia},
	month = sep,
	year = {2018},
	pages = {1800541},
}

@article{guan_lattice_2025,
	title = {Lattice structure musculoskeletal robots: {Harnessing} programmable geometric topology and anisotropy},
	volume = {11},
	issn = {2375-2548},
	shorttitle = {Lattice structure musculoskeletal robots},
	url = {https://www.science.org/doi/10.1126/sciadv.adu9856},
	doi = {10.1126/sciadv.adu9856},
	language = {en},
	number = {29},
	urldate = {2025-08-25},
	journal = {Science Advances},
	author = {Guan, Qinghua and Dai, Benhui and Cheng, Hung Hon and Hughes, Josie},
	month = jul,
	year = {2025},
	pages = {eadu9856},
}

\end{document}